\def\year{2019}\relax
\documentclass[letterpaper]{article} % DO NOT CHANGE THIS
\usepackage{aaai19}  % DO NOT CHANGE THIS
\usepackage{times}  % DO NOT CHANGE THIS
\usepackage{helvet} % DO NOT CHANGE THIS
\usepackage{courier}  % DO NOT CHANGE THIS
\usepackage[hyphens]{url}  % DO NOT CHANGE THIS
\usepackage{graphicx} % DO NOT CHANGE THIS
\usepackage{graphicx}  % DO NOT CHANGE THIS
\usepackage{natbib}
\usepackage{amsmath}
\usepackage{subcaption}
\usepackage{wrapfig}
\usepackage{algorithm}
\usepackage[noend]{algpseudocode}
\usepackage{amssymb}
\usepackage{amsmath}
\usepackage{amsthm}

\usepackage{color}

\newcommand{\Coment}[1]{}

\newcommand{\wprox}{\ensuremath{w}}
\newcommand{\zprox}{\ensuremath{Z}}
\newcommand{\mprox}{\ensuremath{M}}

\newcommand{\rprox}{\ensuremath{\tilde{r}}}
\newcommand{\proxspace}{\ensuremath{\widetilde{\mathcal{R}}}}

\newcommand{\mnor}{\mprox}

\newcommand{\agentmodel}{\ensuremath{\pi(\cdot \mid \rprox)}}
\newcommand{\rstar}{\ensuremath{r^*}}
\newcommand{\rspace}{\ensuremath{\mathcal{R}}}
\newcommand{\poolspace}{\ensuremath\proxspace_{pool}}

\newcommand{\expectation}[2]{\mathbb{E}_{#1} #2}
\newcommand{\piprox}{\ensuremath{\pi(\xi \mid \wprox)}}
\newcommand{\piproxdot}{\ensuremath{\pi(\cdot \mid \wprox)}}

{\begin{list}{$\bullet$}{%
    \setlength{\topsep}{0in}
    \setlength{\partopsep}{0in}
    \setlength{\itemsep}{0in}
    \setlength{\parsep}{0in}
    \setlength{\leftmargin}{3.5em}
    \setlength{\rightmargin}{0in}
    \setlength{\itemindent}{-.1in}
}
}%
{\end{list}
}

\DeclareMathOperator*{\argmax}{argmax}

\newcommand{\prg}[1]{\noindent\textbf{#1. }} 
\PassOptionsToPackage{numbers}{natbib}

\nocopyright
\title{
Active Inverse Reward Design}
\author{
S\"{o}ren Mindermann\thanks{Equal contribution.}, ~~Rohin Shah\footnotemark[1], ~~Adam Gleave, ~~Dylan Hadfield-Menell
\\
\\
UC Berkeley
\\ %If you have multiple authors and multiple affiliations
}
 
\begin{document}

\maketitle
\setcounter{secnumdepth}{2}

\begin{abstract}
Designers of AI agents often iterate on the reward function in a trial-and-error process until they get the desired behavior, but this only guarantees good behavior in the training environment. We propose structuring this process as a series of queries asking the user to compare between different reward functions. Thus we can actively select queries for maximum informativeness about the true reward. In contrast to approaches asking the designer for \emph{optimal} behavior, this allows us to gather additional information by eliciting preferences between \emph{suboptimal} behaviors. %We design two types of queries: discrete queries, where the system designer chooses from a small set of reward functions, and feature queries, where the system queries the designer for weights on a small subset of features.
After each query, we need to update the posterior over the true reward function from observing the proxy reward function chosen by the designer. The recently proposed Inverse Reward Design (IRD) enables this.
%We are able to update the posterior over the true reward function from the observed proxy reward function chosen by the designer after each query thanks to Inverse Reward Design (IRD).
%After each query, we use inverse reward design (IRD) to update the posterior over the true reward function from the observed proxy reward function chosen by the designer. 
Our approach %decreases the uncertainty about the true reward, and 
substantially outperforms IRD in test environments. In particular, it can query the designer about interpretable, linear reward functions and still infer non-linear ones.
\end{abstract}

\section{Introduction}  \label{introduction}

\begin{figure*}[t!]
    \centering
    \begin{subfigure}{0.48\textwidth}
        \includegraphics[width=\textwidth]{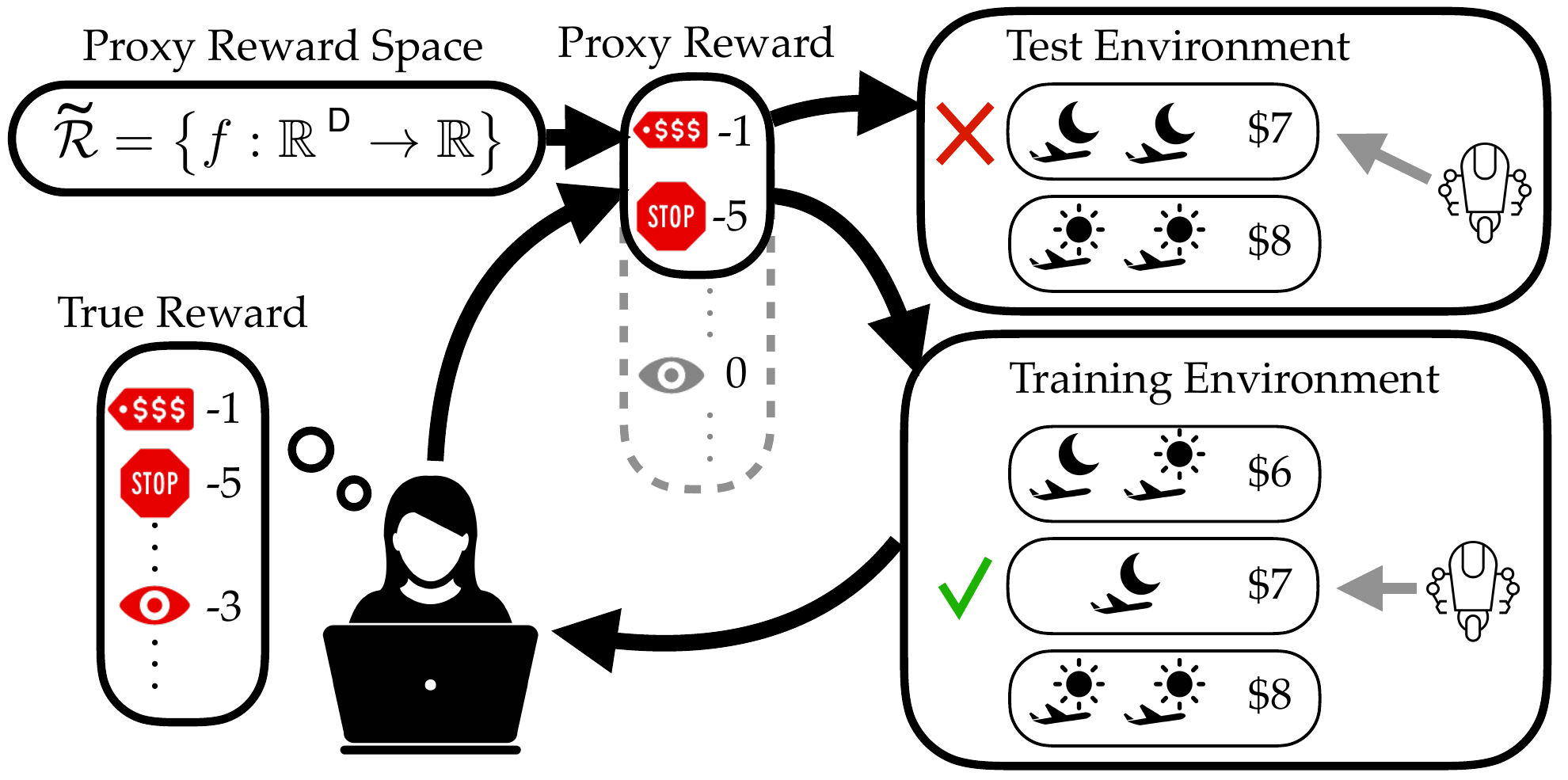}
        \caption{\textbf{Conventional reward design.} Alice iterates on a (proxy) reward function for flight shopping until her agent chooses correctly in the training environment. She wants cheaper flights, but would pay $\$5$ to avoid a layover (stop sign) and $\$3$ to avoid a \emph{red-eye} flight, making the $\$7$ flight optimal. She first assigns a weight of $-1$ to price, but then her agent chooses the suboptimal $\$6$ flight. She realizes that the $\$7$ flight is better because it is nonstop, and so assigns a weight of $-5$ to stops. The robot then selects the optimal $\$7$ flight, and Alice believes her job is done, \emph{not realizing she forgot about red-eye flights}. As a result, her agent buys the suboptimal red-eye flight in the test environment.}
        \label{fig:reward-design}
    \end{subfigure}
    \quad
    \begin{subfigure}{0.48\textwidth}
        \includegraphics[width=0.95\textwidth]{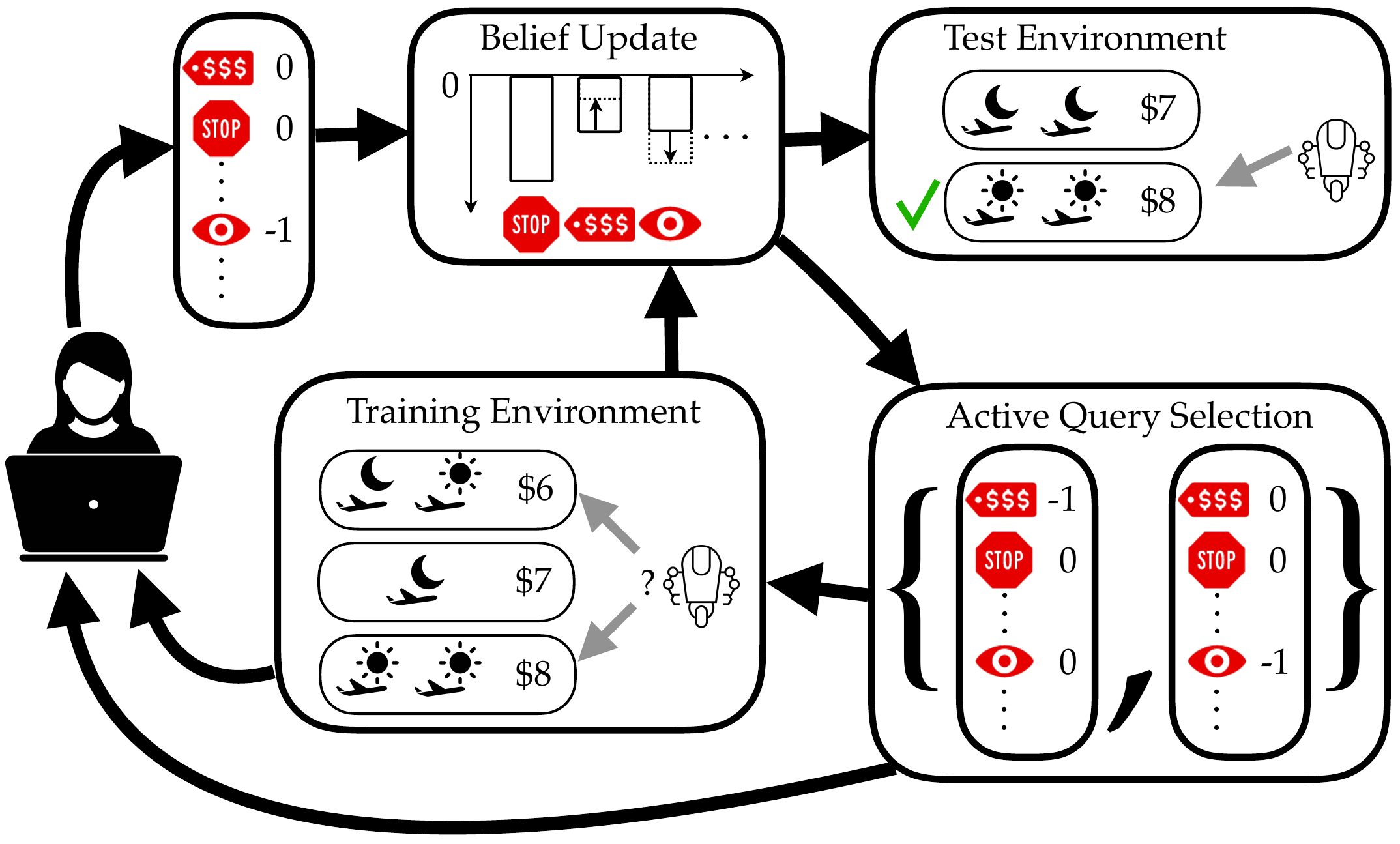}
        \caption{\textbf{AIRD.} We actively select a small set of rewards, known as a \emph{query}, from which Alice must choose the best reward. Suppose we already knew that Alice disprefers layovers, but we do not know Alice's preference about red-eye flights. AIRD designs a query with two reward functions that both assign a weight of $0$ to stops, but differ on price and red-eye flights. Alice chooses the reward that selects the $\$8$ flight, since her other option would choose the worse $\$6$ flight. AIRD can then infer that Alice disprefers red-eye flights, leading the agent to buy the correct flight in the test environment.}
        \label{fig:active_ird}
    \end{subfigure}
    \caption{Illustration of the benefits of Active Inverse Reward Design (AIRD) in a flight shopping environment (which is non-sequential for simplicity). A plane with a sun indicates a daytime leg while a plane with a moon indicates a night time leg. Here, we consider three (of many) features: the cost, the number of intermediate stops, and a binary indicator for red-eye flights. (A flight is red-eye if it has at least one overnight leg.) This simple example is one-off but note that AIRD is for \textit{sequential} tasks.}
    \label{fig:illustration}
    \vspace{-3mm}
\end{figure*}

% One approach for building AI agents breaks the problem into two steps: 1) design a reward function; and 2) write an algorithm to optimize that reward function. In practice, system designers interleave these steps -- after optimizing the designed reward function, they see what behavior it incentivizes, allowing them to design a better reward function.

Reward design, the problem of selecting an appropriate reward function, is both critically important, as it encodes an agent's task, and challenging, as it requires the designer to anticipate all possible behaviors in all possible environments and determine appropriate incentives or penalties for each one. Therefore, it has become a central issue in AI safety and alignment for sequential decision-making \citep{amodei2016concrete}. In practice, designers alternate between two steps: observing how the current behavior is suboptimal, and designing a better reward function. However, this only guarantees that the reward function leads to good behavior \emph{in the training environment}.

Consider for example the flight shopping assistant in Fig.~\ref{fig:reward-design}. Alice iterates on the reward function until her agent picks the \$7 nonstop flight in the training environment. However, she forgot to penalize red-eye flights (flights with an overnight leg), because her agent never incorrectly picked such a flight. The next time her agent has to pick a flight, it incorrectly picks the cheaper red-eye flight.

Inverse Reward Design (IRD) \citep{Hadfield2017} is a recent method that leverages this observation by assuming that Alice's reward function is only a proxy for good behavior \textit{in the training environment}, and performing Bayesian inference over the true reward function. However, this is insufficient: the IRD assumption correctly notes that the only thing we know from Alice's designed reward function is that the nonstop flight is optimal. But Alice's true preference over red-eye flights remains unknown. Indeed, even humans would struggle to tell whether Alice would prefer a cheaper, red-eye flight to an expensive, daytime flight if all they knew was that Alice dislikes layovers!

To improve upon IRD, we need to observe more than what Alice considers \textit{optimal}. Our key insight is that by choosing which reward functions Alice can use, we can elicit Alice's preferences over \textit{suboptimal} behaviors. %, yielding more information than IRD. 
This motivates Active Inverse Reward Design (AIRD), where we break the reward design problem into a series of smaller actively chosen reward design problems, or \emph{queries}, in which Alice is asked to choose from a set of candidate reward functions. After each of Alice's answers, IRD is used to update the belief about the true reward.

Figure~\ref{fig:active_ird} shows what AIRD does in our example. Since we are unsure about Alice's true reward for red-eye flights, AIRD constructs a query that forces her to compare between the suboptimal $\$6$ and $\$8$ flights. Alice chooses the reward that would pick the expensive daytime flight, from which AIRD can infer that the red-eye flight penalty should be higher than the price penalty.

We first study the simplest possible queries: discrete sets of N reward functions, of which Alice tries to choose the best. These queries can already infer the true reward where IRD cannot. However, large discrete queries may be hard to answer. %, if the reward functions are very different. 
We therefore design \emph{feature queries}, which ask Alice to set weights for a few features, given weights for the other features (optimized to be maximally informative).

%In applied reinforcement learning and control, reward designers typically engineer high-level reward features. 
%%This is much easier than engineering a reward function directly from e.g. pixels.
%This helps by letting them reward or penalize just the features that matter most. But they can only make simple reward functions.
%% AIRD captures the benefit without the downside. It can use interpretable, high-level features for the queries and still infer rewards over low-level, complex features.
Even if we use interpretable, high-level features for the queries, we can still infer rewards over low-level features. No other reward inference method we have seen can do this.

Our contributions are as follows: we 1) structure the reward design process as a series of queries that we learn from using IRD; 2) design discrete and feature queries, emphasizing simplicity, usability and informativeness; 3) design algorithms that actively select the most informative queries for Alice to answer; and 4) evaluate this approach using simulated human models in a flight shopping assistant domain and a 2D navigation domain. We find that our method reduces regret at test time compared with vanilla IRD, often fully recovering the true reward function.

% Our results indicate that actively selecting the set of available reward functions is a promising direction for increasing the efficiency and effectiveness of reward design.

\section{Background: Inverse Reward Design}

In \emph{inverse reward design} (IRD)~\citep{Hadfield2017} the goal is to infer a distribution over the true reward function the designer wants to optimize given an observed, but possibly misspecified, proxy reward function. This allows the agent to avoid problems that can arise in test environments due to an incorrect proxy reward function. IRD achieves this by assuming that proxy reward functions are likely to be chosen if they lead to high \textbf{true} utility behavior in the \textbf{training} environment.

We represent the fixed training environment faced by our agent as a Markov decision process without reward, \mnor. The policy is given by $\agentmodel{}$, a distribution over trajectories $\xi$, corresponding to a planning or reinforcement learning algorithm. We assume that the designer selects a proxy reward $\rprox$ from a space of options $\proxspace$ so that $\agentmodel{}$ approximately optimizes a true reward function \rstar{} (which the designer knows implicitly). That is, we assume the designer approximately solves the \emph{reward design problem}~\citep{Singh2009}.

% FORMATTING HACK: avoid awkward line break for \agentmodel{}
In the \textit{inverse} reward design problem, \rprox, \proxspace, and \mbox{\agentmodel{}} are known. The true reward $\rstar\in\rspace$ must be inferred under the IRD assumption that \rprox{} incentivizes approximately optimal behavior in $M$. 

Note that the space of true rewards $\rspace$ need not be the same as the space of proxies $\proxspace$. In this work, we assume that a proxy reward is a linear function of pre-specified features (either hand-coded or learned through some other technique), and so we write it as $\rprox(\xi;~\wprox) = \wprox^\top\phi(\xi)$, where $\phi(\xi)$ are the features of trajectory $\xi$. However, the only assumption we make about the true reward space $\rspace$ is that we can perform Bayesian inference over it: we have a prior and it is feasible to approximately compute a posterior.

\subsection{Observation Model} \label{sec:obsmodel}
 
% The designer models as approximately optimizing the provided proxy reward:
% \begin{equation} \label{eq:agentmodel}
% \piprox \propto \exp \left(\beta_p \wprox^\top\phi(\xi)\right),
% \end{equation}

An optimal designer would choose the proxy reward $\wprox$ that maximizes the expected true value $\expectation{\xi \sim \piproxdot}{\rstar(\xi)}$ in $\mnor$. IRD models the designer as approximately optimal with rationality $\beta$:

\begin{equation} \label{eq:lhood}
P(\wprox \mid \rstar, \proxspace) \propto \exp\left(\beta \ \expectation{\xi \sim \piproxdot}{\rstar(\xi)} \right).
\end{equation}

This model is then inverted to obtain the posterior distribution over true rewards $P(\rstar \mid \wprox, \proxspace)$.

\subsection{Cost of inference} \label{computation}

Computing the likelihood \eqref{eq:lhood} is expensive as it requires calculating the normalization constant $\zprox(\rstar) = \sum_{\wprox\in\proxspace} \exp\left(\beta \ \expectation{\xi \sim \piproxdot}{\rstar(\xi)} \right)$ by summing over all possible proxy rewards and computing a policy for each (AIRD makes this cheaper because it only sums over a small query). We cache a sample of trajectories $\{\xi_i\} \sim\piprox$ for each $\wprox\in\proxspace$ and re-use them to evaluate the likelihood for different potential true reward functions $\rstar\in\rspace$.

% If the reward functions are functions of common features, we can just memoize feature expectations instead.
% \citet{Hadfield2017} explored methods to approximate $\zprox(\rstar)$ by sampling $\wprox$. We instead reduce these computations by making $\proxspace$ much smaller.

Conversely, the normalization constant $Z(\wprox)$ for the \textit{posterior} $P(\rstar{}|\wprox, \proxspace) \propto P(\wprox|\rstar{}, \proxspace)P(\rstar{})$ integrates over $\rstar\in\rspace$ and requires no additional planning. % Approximate inference methods such as MCMC do not compute the normalizer.

% The additional cost is to evaluate the average reward  $\sum r(\xi_i)$ on memoized feature expectations for each $r$. This does not involve planning and can be computed in a single matrix multiplication for finite $\rspace$ and $\proxspace$. Even for nonlinear reward functions, $Z(\wprox)$ is fairly cheap to compute.

\section{Query Design and Selection}

In vanilla IRD, the designer selects an approximately optimal proxy reward $\rprox$ from a large proxy space $\proxspace$. In this work, the designer instead selects the reward function from small actively chosen sets $\proxspace{}_t\subset \proxspace{}$, which we call \emph{queries}. We first outline the criterion used to choose queries, and then design the structure of the queries themselves.

% Reducing the size of the set simplifies the choice for the designer.

\subsection{Active selection criterion}

We maximize the expected information gained about the true reward $\rstar$ given the user's answer $\wprox$~\citep{Houlsby2011}. Let $\mathcal{D}_t=\{\proxspace_{1:t-1}, \wprox_{1:t-1}\}$ denote the previous queries and answers, and $P(\rstar \mid \mathcal{D}_t)$ the current belief over the true reward. We compute the predictive distribution $P(\wprox_t \mid \proxspace_t, \mathcal{D}_t)$ over the user's answer from the IRD observation model $P(\wprox_t \mid\rstar, \proxspace_t)$ in \eqref{eq:lhood} by marginalizing over $\rstar$. 

We write the mutual information, or \textit{expected information gain}, between the random variables $\rstar$ and $\wprox$ as:
\begin{equation} \label{infogain}
\textrm{MI}(\proxspace_t,\mathcal{D}_t) =
\mathcal{H}[\wprox | \proxspace_t, \mathcal{D}_t] -\mathbb{E}_{P(\rstar{}|\mathcal{D}_t)}    \mathcal{H}[\wprox | \proxspace_t, \rstar{}],
\end{equation}
where $\mathcal{H}$ is the entropy $\mathcal{H}(\wprox|\cdot)=-\sum_{\wprox}P(\wprox|\cdot)\log P(\wprox|\cdot)$.

The first term in (\ref{infogain}) is the predictive entropy, the model's uncertainty about Alice's answer. This is a common active learning criterion, and works for active supervised learning \citep{Gal2017}. However, it selects for query-dependent noise in user answers. The second term ensures that in expectation, Alice is not uncertain about her answer.

% Decision-theoretic acquisition functions pick queries that lead to high expected reward after asking. \citet{Akrour2013,Akrour2014} use these for active preference learning. However, these objectives are often computationally intractable and we found little benefit.

% Information-theoretic acquisition functions are typically used in Bayesian active learning settings. These encode minimal assumptions about the test distribution, assuming only that reducing uncertainty about the true reward function leads to higher reward in test environments.

\begin{figure*}[t]
\centering
\begin{subfigure}{0.45\textwidth}
    \begin{algorithmic}[1] % The number tells where the line numbering should start
        \State \textbf{input: }particle representation of current posterior $\hat{P}_{t-1}$; query size $K$
        \State \textbf{output: }informative next query, $\proxspace^*_t$
        \Procedure{DiscreteQuery}{$\hat{P}_{t-1}, K$} 
            % \For{$t \gets 1 \textrm{ to } T$} 
	            \State $\proxspace_{t}^* = \{\wprox_1 \sim \text{Uniform}(\proxspace)\}$
                \For{$i \gets 1 \textrm{ to } K$ }
            		\For{ $\wprox' \textrm{ in } \proxspace$}
            			\State $\text{query} = \proxspace_{t}^* \cup \{\wprox'\}$
            			\State $\mathbb{I}[\wprox'] = \text{MI}(\text{query},\hat{P}_{t-1})$
    				\EndFor
    				\State $\proxspace_t^* = \proxspace_t^* \cup \{ \argmax_{\wprox'} \mathbb{I} \}$
                \EndFor
        % 	    \State $\wprox_t = \textrm{Human}(\proxspace_{ti}^*, \rstar, \beta)  $  
        %     	\State $\mathcal{D}_t = \mathcal{D}_{t-t} \cup \{(\wprox_t,\proxspace_{ti}^*)\}$
        %     	\State $P(\rstar | \mathcal{D}_t) \propto P_{t-1}(\rstar) P(\wprox | \rstar)$
        %     \EndFor
            \State \textbf{return} $\proxspace_{t}^*$
        \EndProcedure
    \end{algorithmic}
    \caption{Greedy discrete query selection.}
    \label{alg:discrete}
\end{subfigure}
\quad
\begin{subfigure}{0.51\textwidth}
    \begin{algorithmic}[1] % The number tells where the line numbering should start
    \State \textbf{input: }particle representation of current posterior $\hat{P}_{t-1}$; feature functions $F$; query size $K$
        \State \textbf{output: }informative next query, $\proxspace^*_t$
        \Procedure{FeatureQuery}{$\hat{P}_{t-1}, F, K$} 
            % \For{$t \gets 1 \textrm{ to } T$} 
            % 	\State $\{r_k\} \sim P(\rstar|\mathcal{D}_{t-1})$
            % 	\State $\hat{P}_{t-1} = \textrm{ EmpiricalDist}(\{r_k\})$
	           % \State $f_1 \sim \text{Uniform}(\proxspace)$
	           % \State $\proxspace_{t1}^* = \{\wprox_1\}$
	            	          %  \Comment{Greedy search initialization}
	            \State $Q^*=\{\}$
                \For{$k \gets 1 \textrm{ to } K$ }
            		\For{ $f$ \textbf{ in } $F$}
                      % \State $\proxspace_t[f] = \textrm{InitFeatQuery}(Q')$
                        \State $\textrm{query} =\textrm{OptimW}( Q^* \cup \{f\}, \hat{P}_{t-1})$
                        \State $\mathbb{I}[f] = \textrm{MI}( \textrm{query},\hat{P}_{t-1})$
    				\EndFor
                    % Problem: Need to preserve fixed weights
    				\State $Q^* = Q^* \cup \{\argmax_f \mathbb{I}\} $
    			%	\State $\proxspace_t^* = \proxspace_t[\argmax_f \mathbb{I}]$
                 \EndFor
            \State \textbf{return} $\proxspace_t^* = \textrm{OptimW}( Q^* , \hat{P}_{t-1})$
        \EndProcedure
    \end{algorithmic}
    \caption{Greedy feature query selection.}
    \label{alg:features}
\end{subfigure}
\caption{Algorithms for query selection. Mutual information MI is calculated with Equation~\ref{infogain}. \textrm{OptimW} optimizes the fixed weights of the feature query, as described in Section~\ref{sec:feature_queries}.}
\vspace{-5mm}
\end{figure*}

\subsection{Discrete queries} \label{sec:discrete}
A natural query type is to pick any finite set of reward functions: $\proxspace_t=\{\wprox_1,\dots,\wprox_K\}$. For small $K$, Alice can inspect the induced policies, so her answer should be nearly optimal.

\prg{Exploiting information about suboptimal proxies} Consider a perfectly rational designer ($\beta \rightarrow \infty$). Vanilla IRD merely learns which proxy reward $\wprox \in \proxspace$ is best, leading to at most $|\proxspace|$ possible outcomes. With discrete queries of size 2, we can compare two arbitrary rewards, 
%allowing us to infer an ordering on $|\proxspace|$. Far more could be inferred because there are up to $|\proxspace|!$ possible \textit{orderings}.
allowing us to learn a complete ordering on $\proxspace$, with up to $|\proxspace|!$ outcomes.
%The designer's answer to \emph{any} query can be perfectly predicted using this ordering.

% This is what allowed us to solve our running example: as shown in Figure~\ref{fig:active_ird} (bottom), Alice's robot chooses a discrete query that has Alice compare between two \emph{suboptimal} choices, cake and eggs, and infers that $F > M$.

\prg{Greedy query selection} Searching over all queries of size $K$ requires a prohibitive $\binom{|\proxspace|}{K}$ evaluations of the expected information gain. We therefore grow queries greedily up to size $K$, requiring only $O(|\proxspace| K)$ evaluations (Algorithm \ref{alg:discrete}). In Section~\ref{sec:discrete-eval}, we empirically find this works as well as a very large random search.

\prg{Proxy pool} Many reward functions lead to the same optimal policy: a major problem for inverse reinforcement learning~\citep{Ng2000}, but an advantage for us. Even if we discard many proxy rewards, most possible behaviors will remain. To this end, we initially uniformly sample a proxy space $\poolspace \subset \proxspace$, and perform active selection from this much smaller subset. We compute belief updates over the arbitrary space $\rspace$, so we can still recover the true reward function $\rstar$.

We precompute trajectory samples $\{\xi_i\}\sim\pi(\cdot \mid w)$ for every proxy reward $\wprox \in \poolspace$, which are needed for the likelihood in \eqref{eq:lhood}. This means that we never need to run planning during query selection or inference, making our method very efficient during designer interaction.

%This facilitates live user interaction in domains where planning is expensive. Our active learning approach (and posterior inference) are thus independent of the difficulty of planning in the training environment once $k$ policies have been pre-computed.

\subsection{Feature queries} \label{sec:feature_queries}

While discrete queries are computationally efficient, they could be hard for Alice to answer. If two reward functions incentivize very different suboptimal behaviors, it can be hard to say which of these is better. When comparing rewards $r_1$ and $r_2$ whose weights differ on many features, some of the differences will favor $r_1$, and some $r_2$, and it may be hard to determine overall which reward is better.

However, the benefit of AIRD is to elicit information from Alice about suboptimal reward functions. How could we query about such rewards, \emph{without} making it hard to answer? We could have Alice tune the weights of a few interpretable features, whose effects are easy to understand.

Since it is hard for Alice to determine the ``true'' weights, the queries should be grounded in particular trajectories. Alice can then consider the effects of feature weights on those trajectories, and not worry about whether her weights are going to lead to bad behavior in edge cases. As a result, we need to provide in the query a \emph{valuation} for all of the features that we are not asking Alice about.

Concretely, suppose we have $D$ features in total, and that weights can range over $\mathbb{R}$. We define a feature query of size $K$ to be a set of $K$ free weights $S \subset \{1,\dots,D\}$ and a valuation of fixed weights $v : \{1,\dots,D\} \setminus S \rightarrow \mathbb{R}$. It corresponds to the set of reward functions given by $\{w\in\mathbb{R}^D:\forall i \in \text{Domain}(v), w_i=v(i) \}$. Alice then specifies weights $w_i$ for $i \in S$ to determine a proxy reward.

In each query, Alice can judge the effects of a few features that are currently most informative. Recent work \citep{Basu2018} shows that this can lead to more efficient learning. Relative to discrete queries, feature queries increase the information received per query, at the cost of computational complexity. They also provide a more intuitive interaction: we could imagine a graphical user interface in which Alice moves sliders for each weight to see the effect on the policy.

\prg{Discretization} Exactly evaluating the expected information gain is only tractable for finite queries. We therefore discretize the free (but not the fixed) weights.

\prg{Feature query selection} There are two variables to optimize over: which features are free, and the values of the fixed features. We select the $K$ free features greedily, similarly to discrete queries, as shown in Algorithm~\ref{alg:features}. 

To tune the fixed weights, we need to optimize over a continuous space $\mathbb{R}^{D-K}$. 
We apply gradient descent using a differentiable implementation of soft value iteration based off of \citet{Tamar2016}. Due to problems with local maxima, we used a small random search over $\mathbb{R}^{D-K}$ to find a good initialization, improving results considerably. Random search by itself works reasonably well and can be used when differentiable planning algorithms are not available. Feature queries cost more compute but even with the most expensive settings it was a few seconds per query on one GPU.

\prg{Scalability} Feature queries are particularly useful given interpretable features; yet we would like to infer reward functions on  low-level features (e.g. pixels) with complex interactions. This is feasible, since the inference over the true reward space depends only on the \emph{behavior} incentivized by the proxy reward functions and not on the rewards themselves. In Section~\ref{sec:nonlinear-eval} we find that it is sufficient to use only queries about linear rewards to infer a true reward function with hundreds of interaction terms.

\section{Evaluation}

Our primary metric is the test environment regret obtained when we plan using the posterior mean reward $\bar{r}=\mathbb{E}[\rstar{}|\mathcal{D}]$ across a set of unseen test environments. We supplement this with another metric, the entropy of the agent's belief $\mathcal{H}[\rstar]$, which measures how uncertain the agent is about the true reward. We selected $20$ queries per experiment and reported the two measures after each query. Human input is simulated with the likelihood \eqref{eq:lhood}.

% We seek to answer the following questions: (1) do many small queries help more than a single large query, as hypothesized in Section~\ref{sec:discrete}, (2) how much does active selection improve upon random selection, (3) does the heuristic of greedy selection sacrifice substantial performance, (4) for feature queries, how much does free feature selection and valuation optimization help, (5) is it feasible to infer low-level, complex rewards using queries on high-level, interpretable features, and (6) which queries are most sample efficient?

\prg{Environments}
Active IRD is performed on one training environment, and evaluated on many random test environments (all share $\rstar$). We used two domains. The flight shopping domain is a simple one-step decision problem, with one state and $100$ actions (flights). Each flight is described by $20$ features determined by \textit{iid.} draws from a Gaussian distribution. Conceptually, a robot is trained with one set of flights to select unseen flights in other such sets. The 2D navigation task is a featurized $10\times10$ gridworld called a ``Chilly World'' with random walls and $20$ objects in random positions. $20$ features are given by the Euclidean distances to each object. $\rstar$ describes the `temperatures' of each object: hot objects should be approached and cold ones avoided. We compute the policy with $20$ steps of soft value iteration, which is differentiable. Further parameters are in appendix A.

\prg{True reward space $\rspace$} While in principle our method can be applied to any $\rspace$ amenable to Bayesian inference, for computational efficiency we consider a finite space of true reward functions with $|\rspace| = 10^6$ unless otherwise specified. As a result, we can compute the exact distribution over true rewards. This allows us to evaluate the effect of our queries without worrying about variance in the results arising from randomness in approximate inference algorithms.

\begin{figure*}[t]
    \begin{subfigure}{0.48\textwidth}
        \centering
        \includegraphics[width=\textwidth]{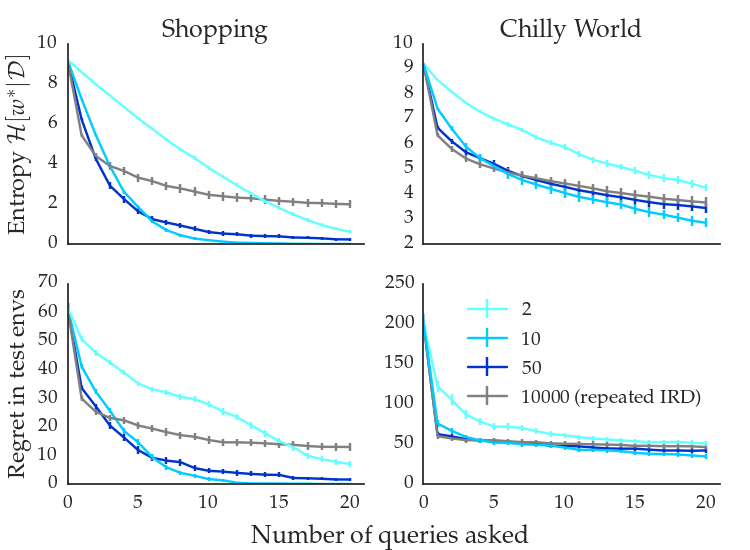}
        \caption{Random discrete queries of sizes 2, 10, 50, and repeated exact full IRD with a proxy reward space of size 10000. Larger queries lead to faster initial learning, but can hurt final performance.}
        \label{fig:outperform}
    \end{subfigure}
    \quad
    \begin{subfigure}{0.48\textwidth}
        \centering
        \includegraphics[width=\textwidth]{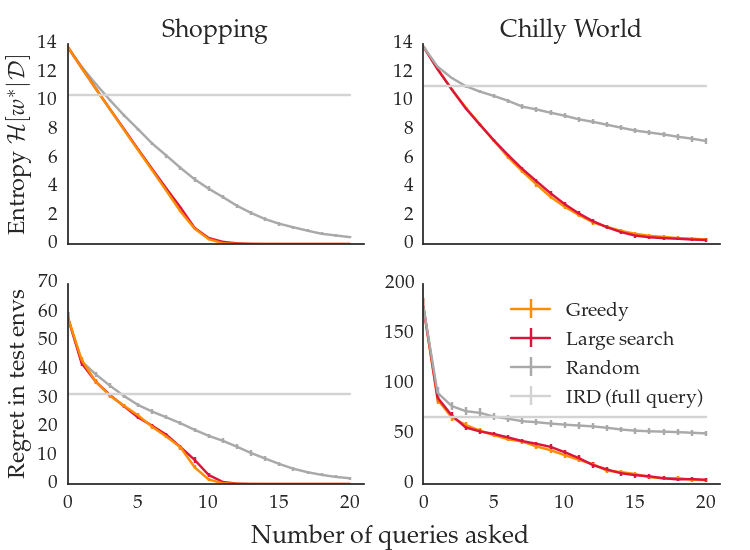}
        \caption{Discrete query selection methods (query size 5) and full IRD. Note that the cheap greedy selection matches the expensive search and IRD remains too uncertain to generalize well.}
        \label{fig:discrete}
    \end{subfigure}
    \vspace{-2mm}
    \caption{Results for discrete queries.}
    \vspace{-5mm}
\end{figure*}

\subsection{Benefits of small queries} 

In Section~\ref{sec:discrete} we hypothesized that smaller queries allow us to compare \emph{suboptimal} behaviors, which vanilla IRD cannot do. To test this, we compare the performance of \textit{randomly} chosen discrete queries of various sizes. Note that full IRD is equivalent to a $|\proxspace|$ query size. In this experiment only, we used the maximal proxy space $\proxspace=\rspace$, and reduced $|\rspace|$ to $10^4$ to make exact IRD feasible. IRD was run $20$ times to show its convergence behavior, although it would normally be run only once.

Figure~\ref{fig:outperform} shows that using a smaller query attains better generalization performance than full IRD after as few as five queries, validating our hypothesis. Note that performance on the first query increases with query size: a small query size only helps after a few queries, when it becomes necessary to compare between suboptimal behaviors since the optimal behavior in the training environment has been mostly found.

\subsection{Discrete query selection} \label{sec:discrete-eval}

We next turn to greedy discrete query selection (Algorithm~\ref{alg:discrete}). To evaluate how useful active selection is, we compare to a baseline of random query selection. To evaluate whether the greedy heuristic sacrifices performance, we would like to compare to a baseline that searches the entire space of discrete queries. However, this is computationally infeasible, and so we compare against a large search over $10^4$ random queries (which is still much slower than greedy selection). $|\poolspace|$ was set to $100$.%, more than enough to distinguish between $10^6$ potential true rewards.

Figure~\ref{fig:discrete} shows that active selection substantially outperforms random queries and full IRD. Active selection becomes more important over time, likely because a random query is unlikely to target the small amount of remaining uncertainty at later stages. Moreover, greedy query selection matches a large search over $10^4$ random queries.%, confirming previous empirical results showing greedy algorithms are approximately optimal for information gain \citep{Sharma2015}.

\begin{figure*}[t]
    \begin{subfigure}[t]{0.48\textwidth}
        \centering
        \includegraphics[width=\textwidth]{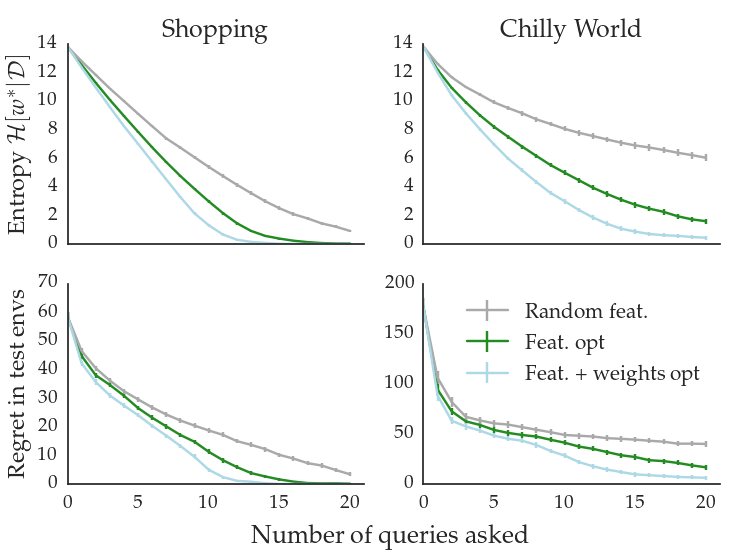}
        \caption{Feature query selection methods with 1 free feature, comparing 1) unoptimized random feature query 2) free feature actively selected 3) additionally, fixed weights optimized.}
        \label{fig:feature-queries-1}
    \end{subfigure}
    \quad
    \begin{subfigure}[t]{0.48\textwidth}
        \centering
        \includegraphics[width=\textwidth]{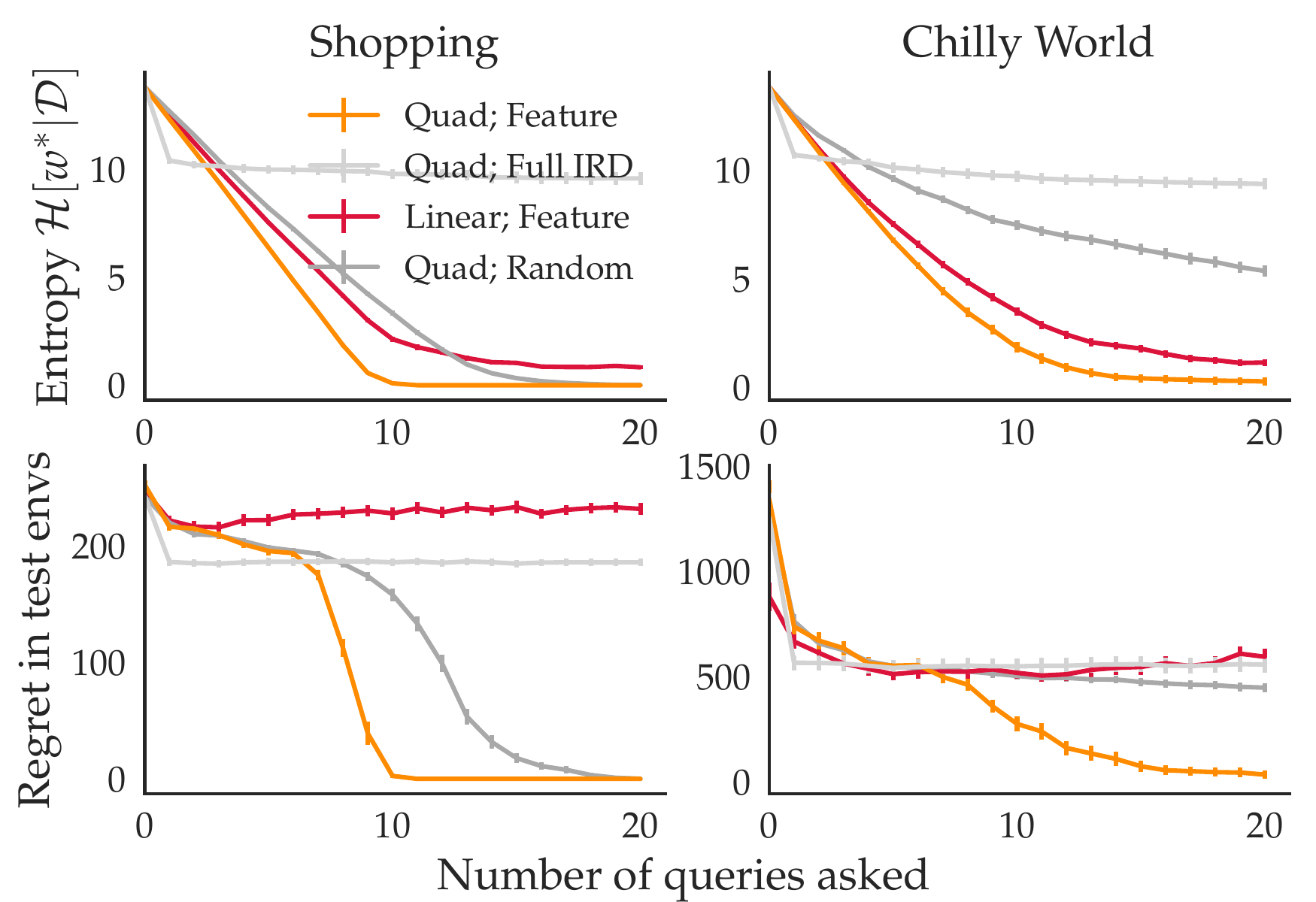}
        \caption{Nonlinear true reward. Comparison between size 1 feature queries with linear proxy rewards and a linear vs quadratic true reward space vs randomly chosen feature queries vs repeated full IRD. % HARD to read!
        % Note that the true reward is always quadratic, so the case with a linear true reward space involves an incorrect assumption, and so performs very badly on test regret.
        }
        \label{fig:nonlinear}
    \end{subfigure}
    \vspace{-2mm}
    \caption{Results for feature queries.}
    \vspace{-1mm}
\end{figure*}

% \begin{wrapfigure}{R}{0.48\textwidth}
\begin{figure}
    \vspace{-4mm}
    \centering
    \includegraphics[width=0.48\textwidth]{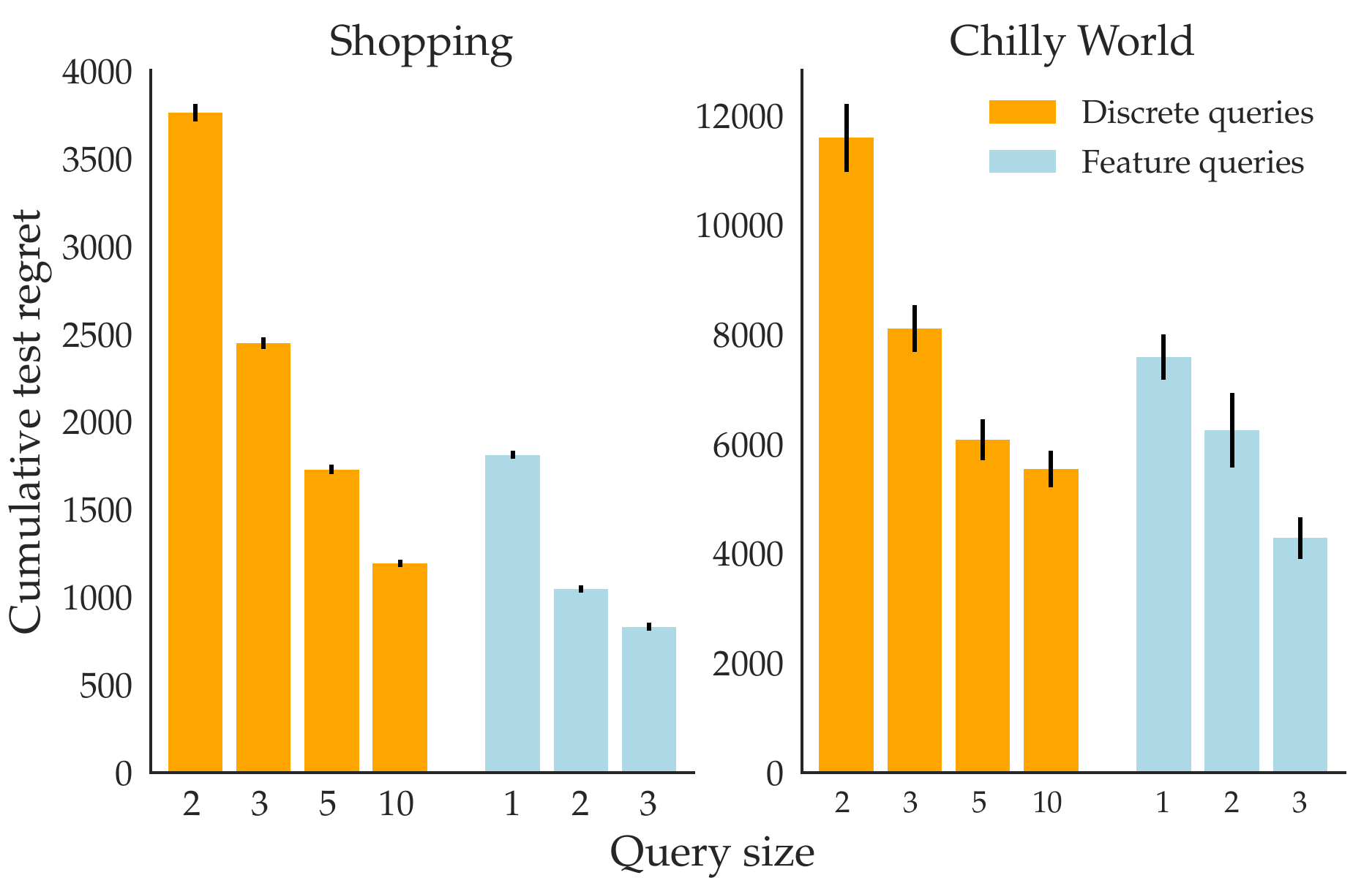}
    \caption{Cumulative test regret for discrete and feature queries of different sizes using the best-performing selection methods to infer non-linear true rewards with linear proxies.}
    \label{fig:sample-efficiency}
    \vspace{-4mm}
\end{figure}
% \end{wrapfigure}

\subsection{Feature query selection}

For feature queries, we would like to evaluate how useful it is to optimize each part of the query. So, we compare among three alternatives: (1) randomly choosing free features, (2) actively selecting free features, and (3) actively selecting free features and optimizing the valuation of fixed features. In both (1) and (2), the fixed weights are set to $0$. Note that feature queries are not restricted to asking about rewards in $\poolspace$: they can ask about any reward in $\mathbb{R}^D$.

Figure~\ref{fig:feature-queries-1} compares these alternatives for feature queries with $1$ free feature. We find that both parts of the query should ideally be optimized. However, we can save computation and still get good results by just selecting free features.

\subsection{Inferring complex, low-level rewards} \label{sec:nonlinear-eval}

In Section~\ref{sec:feature_queries}, we argued that AIRD can ask queries over interpretable features, while performing inference over a low-level space such as neural nets. We can test this in our environments by using AIRD to infer a high-dimensional \emph{quadratic} reward over features, while constraining the queries to only contain \emph{linear} rewards over features. % Active IRD still achieves good results (Figure \ref{fig:nonlinear}).

Figure~\ref{fig:nonlinear} shows that feature queries work well even in this setting. Note that performing inference under the incorrect assumption that the true reward is linear leads to low entropy of our distribution over true rewards but high test regret. This indicates that accurately modeling the nonlinear part of the reward is crucial. Feature queries still outperform both random queries and full IRD, suggesting that both actively selecting queries and eliciting information about suboptimal rewards are necessary for the best performance.

\subsection{Sample efficiency of queries}

A sample efficient algorithm will have lower test regret earlier in the process. Thus, we measure the \emph{cumulative} test regret, that is, the sum of the test regrets after each query. We compare discrete and feature queries over linear proxy rewards. The true rewards are nonlinear.

Figure~\ref{fig:sample-efficiency} shows that for discrete queries, larger query sizes substantially reduce cumulative test regret. We unrealistically assumed here that larger queries do not decrease the quality of human answers. Feature queries are effective even when only tuning a single feature at a time.

\section{Discussion} \label{sec:discussion}

\prg{Query size} Large queries, including full IRD, consistently perform better at first. But Figure~\ref{fig:outperform} shows that relatively small queries can converge to much lower test regret. This is because a large query with all options can extract a lot of information initially but subsequently, a targeted query is necessary to learn from suboptimal rewards. In our running example in Figure~\ref{fig:illustration}, a small query was needed to identify that Alice dislikes redeye flights. Future research could investigate the utility of varying the query size from large to small or fine-tuning a non-active method with an active one.

% All figures show that large queries
% Figure~\ref{fig:outperform} and \ref{fig:sample-efficiency} show that large queries work best initially

\prg{Rewards vs. trajectories} Fundamentally, preference inference methods are useful because human evaluations of behavior are much better than human reward design. The key idea of IRD \citep{Hadfield2017} is to apply this insight to the reward design process itself: it infers a posterior over reward functions assuming that the designed reward function is itself only conveying information about optimal behavior in the training environment. For this reason, we ensured that Alice could answer any of our queries by looking at the trajectories induced by the rewards. (That is, we include fixed weights for feature queries, instead of simply asking Alice for the trade-off between the free features without considering other features.) One might wonder why we would go through all of this trouble instead of just directly using trajectory queries. However, reward queries have several advantages:
\vspace{-1.5mm}

\begin{itemize}
    %\item Reward space is typically smaller and more structured than trajectory space, so it can be easier to find a good reward query, especially with engineered, interpretable features. 
    %\item Reward space is typically smaller and more structured than trajectory space, so it can be easier to find a good reward query. In particular, reward designers routinely engineer features (e.g. distances, velocities, accelerations) since learning reward features is hard given limited human data. We optimize reward queries over the associated weight space.
    \item Reward space is typically (much) smaller and more structured than trajectory space, so it can be easier to find a good reward query. For example, reward spaces often use feature engineering or outputs of a pose-estimation model \citep{andrychowicz2018learning}. Trajectory methods do not use this information.% (e.g. distances, velocities, accelerations) and reward queries be optimized over the weight space.% Since learning reward features is hard given limited human data. We optimize reward queries over the associated weight space.
    \vspace{-1.7mm}
    \item With feature queries with interpretable features, Alice may tune weights directly without considering trajectories if she so desires.
    \vspace{-1.7mm}
    \item Feature queries use the structure of reward space to ask about hundreds of rewards at once without overwhelming Alice. This can't be done with trajectory queries.
    \vspace{-1.7mm}
    \item Even if Alice evaluates reward functions via induced trajectories, the trajectories are optimal for some reward, so are easier to understand than random trajectories.
\end{itemize}

\section{Related work}

A variety of approaches for learning reward functions exist. \emph{Inverse reinforcement learning} (IRL) \citep{Ng2000,Ramachandran2007,Ziebart2008} observes demonstrations of roughly optimal behavior, and infers a reward function that explains it. Reward functions have also been learned from \emph{expert ratings} \citep{Daniel2014} and \emph{human reinforcement} \citep{Knox2009,Warnell2017}. Similarly, several approaches apply active selection to %environments~\citep{amin2017repeated} or 
trajectory comparisons~\citep{Sadigh2017, cui2018active}.

There is a vast number of preference learning methods for \textit{non}-sequential problems \citep{furnkranz2010preference}. We used such a problem for illustration, but AIRD (and IRD) is generally overkill here.

Methods that learn reward functions from preferences, surveyed in \citet{Wirth2017}, are particularly relevant to our work. \citet{Christiano2017} learn a reward function from preferences over pairs of trajectories, by sampling trajectories from a learned policy and querying the user about pairs with high uncertainty. A similar setup is used in \citet{Wirth2016} and \citet{Akrour2012} based around other policy optimization methods. It is also possible to learn reward functions from preferences on actions \citep{Furnkranz2012} and states \citep{Runarsson2014}. Many of these methods select from already encountered trajectories, which will be guesses at \emph{optimal} training behavior. In contrast, AIRD can ask about suboptimal rewards.

Our work is most similar to \citet{Sadigh2017}, which can find queries by gradient-based optimization in the trajectory space (in continuous environments). %Their objective is expected volume removed from the hypothesis space by the query, which is similar to our method of optimizing for information gain. 
We actively select reward functions rather than trajectories (see Section~\ref{sec:discussion}).

%Moreover, since we can maintain a well-calibrated distribution over true rewards, we know how far we are from obtaining the true reward function $\rstar$ (as long as $\rstar \in \rspace$). In some cases, we can exactly recover $\rstar$, guaranteeing generalization to new environments.

\section{Limitations and Future Work}

\prg{Summary} Inverse reward design (IRD) learns from the final proxy reward chosen by the designer. In contrast, active IRD structures the reward design process as a series of simpler reward design queries, and uses the IRD update after each query to learn from intermediate choices as well. During this process, we can ask the designer to compare between \emph{suboptimal} rewards, yielding information not available with vanilla IRD. We designed two types of queries that trade off between usability, computational efficiency and sample efficiency. We demonstrated that this leads to better identification of the correct reward and reduced regret in novel environments, even when we must infer a low-level reward function using queries on high-level features.

\prg{Realistic environments} Our primary contribution is a conceptual investigation of a novel approach to learning reward functions. As a result, we have focused on simple environments which do not require a huge engineering effort to get results. As we saw in Section~\ref{sec:nonlinear-eval}, there is no \emph{conceptual} difficulty with realistic environments with non-linear rewards. Of course, inference in more complex spaces poses challenges that we hope to explore in future work. We would particularly like to test the hypothesis that utilizing the engineered reward features typically used in deep RL allows for more efficient queries than trajectory methods (section \ref{sec:discussion}).

\prg{Unseen features} Another limitation is that when some feature is absent in the training environment, as in the Lava world example in \citet{Hadfield2017}, we cannot infer its reward weight. We intend to try mitigating this through active design of environments, e.g. as in \cite{amin2017repeated}. 

\prg{User studies} While our evaluation established the performance benefits of active IRD, this was under a simulated human model. We would like to perform user studies to test the accuracy of real designers on various types of queries. This would also help test our hypothesis that users are more accurate at picking from a small set than a large proxy space.

\bibliographystyle{apalike}
\bibliography{references}

\begin{thebibliography}{}

\bibitem[Akrour et~al., 2012]{Akrour2012}
Akrour, R., Schoenauer, M., and Sebag, M. (2012).
\newblock {APRIL}: Active preference learning-based reinforcement learning.
\newblock In {\em ECMLPKDD}, pages 116--131.

\bibitem[Amin et~al., 2017]{amin2017repeated}
Amin, K., Jiang, N., and Singh, S. (2017).
\newblock Repeated inverse reinforcement learning.
\newblock In {\em NIPS}, pages 1815--1824.

\bibitem[Amodei et~al., 2016]{amodei2016concrete}
Amodei, D., Olah, C., Steinhardt, J., Christiano, P., Schulman, J., and
  Man{\'e}, D. (2016).
\newblock Concrete problems in ai safety.
\newblock {\em arXiv preprint arXiv:1606.06565}.

\bibitem[Andrychowicz et~al., 2018]{andrychowicz2018learning}
Andrychowicz, M., Baker, B., Chociej, M., Jozefowicz, R., McGrew, B., Pachocki,
  J., Petron, A., Plappert, M., Powell, G., Ray, A., et~al. (2018).
\newblock Learning dexterous in-hand manipulation.
\newblock {\em arXiv preprint arXiv:1808.00177}.

\bibitem[Basu et~al., 2018]{Basu2018}
Basu, C., Singhal, M., and Dragan, A.~D. (2018).
\newblock Learning from richer human guidance: Augmenting comparison-based
  learning with feature queries.
\newblock In {\em HRI}, pages 132--140.

\bibitem[Christiano et~al., 2017]{Christiano2017}
Christiano, P.~F., Leike, J., Brown, T., Martic, M., Legg, S., and Amodei, D.
  (2017).
\newblock Deep reinforcement learning from human preferences.
\newblock In {\em NIPS}, pages 4302--4310.

\bibitem[Cui and Niekum, 2018]{cui2018active}
Cui, Y. and Niekum, S. (2018).
\newblock Active reward learning from critiques.
\newblock In {\em ICRA}, pages 6907--6914. IEEE.

\bibitem[Daniel et~al., 2014]{Daniel2014}
Daniel, C., Viering, M., Metz, J., Kroemer, O., and Peters, J. (2014).
\newblock Active reward learning.
\newblock In {\em RSS}.

\bibitem[F{\"u}rnkranz and H{\"u}llermeier, 2010]{furnkranz2010preference}
F{\"u}rnkranz, J. and H{\"u}llermeier, E. (2010).
\newblock {\em Preference learning}.
\newblock Springer.

\bibitem[F{\"u}rnkranz et~al., 2012]{Furnkranz2012}
F{\"u}rnkranz, J., H{\"u}llermeier, E., Cheng, W., and Park, S.-H. (2012).
\newblock Preference-based reinforcement learning: a formal framework and a
  policy iteration algorithm.
\newblock {\em Machine Learning}, 89(1):123--156.

\bibitem[Gal et~al., 2017]{Gal2017}
Gal, Y., Islam, R., and Ghahramani, Z. (2017).
\newblock Deep {Bayesian} active learning with image data.
\newblock In {\em ICML}.

\bibitem[Hadfield-Menell et~al., 2017]{Hadfield2017}
Hadfield-Menell, D., Milli, S., Abbeel, P., Russell, S.~J., and Dragan, A.
  (2017).
\newblock Inverse reward design.
\newblock In {\em NIPS}.

\bibitem[Houlsby et~al., 2011]{Houlsby2011}
Houlsby, N., Huszar, F., Ghahramani, Z., and Lengyel, M. (2011).
\newblock Bayesian active learning for classification and preference learning.
\newblock {\em CoRR}, abs/1112.5745.

\bibitem[Knox and Stone, 2009]{Knox2009}
Knox, W.~B. and Stone, P. (2009).
\newblock Interactively shaping agents via human reinforcement: The {TAMER}
  framework.
\newblock In {\em KCAP}, pages 9--16. ACM.

\bibitem[Ng and Russell, 2000]{Ng2000}
Ng, A.~Y. and Russell, S.~J. (2000).
\newblock Algorithms for inverse reinforcement learning.
\newblock In {\em ICML}.

\bibitem[Ramachandran and Amir, 2007]{Ramachandran2007}
Ramachandran, D. and Amir, E. (2007).
\newblock Bayesian inverse reinforcement learning.
\newblock In {\em IJCAI}.

\bibitem[Runarsson and Lucas, 2014]{Runarsson2014}
Runarsson, T.~P. and Lucas, S.~M. (2014).
\newblock Preference learning for move prediction and evaluation function
  approximation in othello.
\newblock {\em IEEE Transactions on Computational Intelligence and AI in
  Games}, 6(3):300--313.

\bibitem[Sadigh et~al., 2017]{Sadigh2017}
Sadigh, D., Dragan, A., Sastry, S., and Seshia, S.~A. (2017).
\newblock Active preference-based learning of reward functions.
\newblock In {\em RSS}.

\bibitem[Singh et~al., 2009]{Singh2009}
Singh, S., Lewis, R.~L., and Barto, A.~G. (2009).
\newblock Where do rewards come from?
\newblock In {\em CogSci}, pages 2601--2606.

\bibitem[Tamar et~al., 2016]{Tamar2016}
Tamar, A., WU, Y., Thomas, G., Levine, S., and Abbeel, P. (2016).
\newblock Value iteration networks.
\newblock In {\em NIPS}, pages 2154--2162.

\bibitem[Warnell et~al., 2017]{Warnell2017}
Warnell, G., Waytowich, N.~R., Lawhern, V., and Stone, P. (2017).
\newblock Deep {TAMER:} interactive agent shaping in high-dimensional state
  spaces.
\newblock {\em CoRR}, abs/1709.10163.

\bibitem[Wirth et~al., 2017]{Wirth2017}
Wirth, C., Akrour, R., Neumann, G., and F\"{u}rnkranz, J. (2017).
\newblock A survey of preference-based reinforcement learning methods.
\newblock {\em JMLR}, 18(1):4945--4990.

\bibitem[Wirth et~al., 2016]{Wirth2016}
Wirth, C., Furnkranz, J., Neumann, G., et~al. (2016).
\newblock Model-free preference-based reinforcement learning.
\newblock In {\em AAAI}, pages 2222--2228.

\bibitem[Ziebart et~al., 2008]{Ziebart2008}
Ziebart, B.~D., Maas, A.~L., Bagnell, J.~A., and Dey, A.~K. (2008).
\newblock Maximum entropy inverse reinforcement learning.
\newblock In {\em AAAI}, volume~8, pages 1433--1438. Chicago, IL, USA.

\end{thebibliography}

\appendix
\clearpage
{

\section{Experiment details}

To aid reproduction, we list hyperparameters that we have not specified yet below. We will also make our code base available when publishing. 

\subsection{Detailed environment descriptions}
The flight shopping domain has 100 states, which correspond to flights. Taking action $i$ corresponds to going to state $i$ and picking flight $i$. After one action, a reward is collected and the episode ends. Each flight has 20 features. When each (training or test) environment is generated, each flight is assigned a 20-dimensional feature vector from a standard Gaussian, which describes how the flight scores on each feature.

The Chilly World domain has 10x10 states in a grid. The agent starts in an arbitrary position and can move in all four directions, but not diagonally. Each training and test environment is generated as follows: First, a set of 20 'objects' are (uniformly) randomly placed on the grid. Then, each free call is assigned a 'wall' with probability 0.3. The agent cannot go through walls and so has to take occasional detours (walls are a non-essential property of this environment). Each state has a feature vector which is given by the negative Euclidean distances to each object. Thus, being close to one or more 'hot' objects (those with positive reward) and away from 'cold' ones is ideal. Many policies are possible because the agent may have to take a somewhat convoluted path to get to the ideal state (or sometimes it should not go to the ideal state because it would have to pass through penalized areas). In a test environment, the same objects will be in different places, so it is helpful to know the weights of all objects.

\subsection{Algorithm parameters} 

Most parameters affect results moderately or barely but the assumed human accuracy $\beta$ may require tuning to application specific data like in most preference inference methods. A too low value is inefficient and a high value can lead to overfitting.

\begin{itemize}
\item Learning rate: 20
\item Gradient descent steps and random searches for feature queries: both 20
\item During query search, free features are discretized to take 9, 5 or 3 values respectively for 1, 2 or 3 free features (larger values had little benefit). Through combination, this leads to e.g. $3^3$ reward functions in a query given 3 free features. Any finer discretization can be used for human input.
\item Weight distribution for random search: Unit co-variance normal.
\item Posterior samples to estimate information gain: 5000 (much smaller works as well)
\item $\beta$ for simulated human and human accuracy assumed in the inference: 0.5 (this describes the assumed accuracy of the human and their actual accuracy used for simulated answers.) 
\item Size of large random search baseline: 10000 queries
\item Number of value iterations for navigation domain: 15 (the shopping domain is a zero-uncertainty contextual bandit problem so soft value iteration can be reduced to a softmax).
\item Temperature parameter for soft value iteration: 0.5

\end{itemize}

\subsection{Environment parameters}

\begin{itemize}
\item Discount parameter $\gamma$: 1
\item Number of test environments: 100
\item Time horizon: 20 steps (induced by 20 steps of value iteration)

\end{itemize}

\subsection{Reward parameters}
The true reward space and the proxy reward space (for discrete queries) consist of vectors ranging from -9 to 9. The discretization is done by uniform sampling and the prior is uniform over the samples. The true reward is chosen uniformly from this space. The free features for feature queries range in the same interval of -9 to 9.

\subsection{Details on evaluation metric}
As previously stated, the posterior mean reward is used for planning in test environments. This mean only exists because all our reward functions (even non-linear ones) can be represented as vectors which form inner products with the same set of features. A different evaluation method would have to be chosen if the functions in the true reward space don't share the same features as their input. However, we could in principle perform inference over a neural network space and constrain all but the last layer to be shared between reward functions, so that each reward function is identified by a vector. Alternatively, the maximum-a-posteriori (MAP) reward could be used.

}

\end{document}